\documentclass[letterpaper, 10 pt, conference]{ieeeconf}  % Comment this line out if you need a4paper

\IEEEoverridecommandlockouts                              % This command is only needed if 
\makeatletter
\let\@IEEEhyperreftrue\relax
\makeatother

\usepackage{cite}
\usepackage{amsmath,amssymb,amsfonts}
\usepackage{algorithmic}
\usepackage{graphicx}
\usepackage{textcomp}
\usepackage{xcolor}
\usepackage{makecell}
\usepackage{array}
\usepackage{booktabs}
\usepackage{multirow}
\usepackage{tabularx}

\usepackage{pifont,xcolor}

\newcommand{\cmark}{\textcolor{green}{\ding{51}}}   % 绿色 ✅
\newcommand{\xmark}{\textcolor{red}{\ding{55}}}     % 红色 ❌
\newcommand{\cmid}{\textcolor{orange}{\ding{51}}}   % 橙色 ✅
\definecolor{lightblue}{rgb}{0.7,0.9,1}

\usepackage{hyperref}

\title{\LARGE \bf
nuPlan-R: A Closed-Loop Planning Benchmark for Autonomous Driving via Reactive Multi-Agent Simulation
}

\author{Mingxing Peng, Ruoyu Yao, Xusen Guo, and Jun Ma, \textit{Senior Member, IEEE}% <-this % stops a space
\thanks{Mingxing Peng and Xusen Guo are with the Intelligent Transportation Thrust, The Hong Kong University of Science and Technology (Guangzhou), Guangzhou 511453, China (e-mail: mpeng060@connect.hkust-gz.edu.cn; xguo796@connect.hkust-gz.edu.cn).}%
\thanks{Ruoyu Yao is with the Robotics and Autonomous Systems Thrust, The Hong Kong University of Science and Technology (Guangzhou), Guangzhou 511453, China (e-mail: ryao092@connect.hkust-gz.edu.cn).}%
\thanks{Jun Ma is with the Robotics and Autonomous Systems Thrust, The Hong Kong University of Science and Technology (Guangzhou), Guangzhou 511453, China, and also with the Division of Emerging Interdisciplinary Areas, The Hong Kong University of Science and Technology, Hong Kong SAR, China (e-mail: jun.ma@ust.hk). }%
}

\begin{document}

\maketitle
\thispagestyle{empty}
\pagestyle{empty}

%%%%%%%%%%%%%%%%%%%%%%%%%%%%%%%%%%%%%%%%%%%%%%%%%%%%%%%%%%%%%%%%%%%%%%%%%%%%%%%%
\begin{abstract}
Recent advances in closed-loop planning benchmarks have significantly improved the evaluation of autonomous vehicles. However, existing benchmarks still rely on rule-based reactive agents such as the Intelligent Driver Model (IDM), which lack behavioral diversity and fail to capture realistic human interactions, leading to oversimplified traffic dynamics. To address these limitations, we present nuPlan-R, a new reactive closed-loop planning benchmark that integrates learning-based reactive multi-agent simulation into the nuPlan framework. Our benchmark replaces the rule-based IDM agents with noise-decoupled diffusion-based reactive agents and introduces an interaction-aware agent selection mechanism to ensure both realism and computational efficiency. Furthermore, we extend the benchmark with two additional metrics to enable a more comprehensive assessment of planning performance. Extensive experiments demonstrate that our reactive agent model produces more realistic, diverse, and human-like traffic behaviors, leading to a benchmark environment that better reflects real-world interactive driving. We further reimplement a collection of rule-based, learning-based, and hybrid planning approaches within our nuPlan-R benchmark, providing a clearer reflection of planner performance in complex interactive scenarios and better highlighting the advantages of learning-based planners in handling complex and dynamic scenarios. These results establish nuPlan-R as a new standard for fair, reactive, and realistic closed-loop planning evaluation. We will open-source the code for the new benchmark.
\end{abstract}

%%%%%%%%%%%%%%%%%%%%%%%%%%%%%%%%%%%%%%%%%%%%%%%%%%%%%%%%%%%%%%%%%%%%%%%%%%%%%%%%
\section{INTRODUCTION}
Reliable evaluation of autonomous driving systems requires simulation environments that faithfully reflect the complexity of real-world traffic. Early benchmarks such as Waymo~\cite{ettinger2021large} and nuScenes~\cite{caesar2020nuscenes} focus on open-loop motion forecasting, which evaluates trajectory accuracy without considering the ego vehicle’s influence on its surroundings. As a result, they cannot effectively assess planning algorithms, as real-world driving inherently involves continuous feedback and multi-agent interactions. Subsequent closed-loop benchmarks, including nuPlan~\cite{caesar2021nuplan} and Waymax~\cite{gulino2023waymax}, enable long-horizon simulation of planners within interactive environments. However, these benchmarks still rely on rule-based reactive agents, typically modeled by the Intelligent Driver Model (IDM). IDM simplifies traffic interactions through fixed heuristics with limited responsiveness, leading to rigid behaviors. For example, in complex scenarios such as junctions, IDM often behave non-cooperatively, producing unrealistic and unsafe interactions that deviate from natural human driving. Consequently, current benchmarks struggle to assess how planners perform in complex, dynamic, and interactive scenarios.

Recently, learning-based multi-agent traffic simulation has made notable progress toward improving realism and stability in closed-loop settings. Diffusion Forcing~\cite{chen2024diffusion} presents a unified perspective that bridges autoregressive prediction and diffusion modeling through a partial-masking mechanism, achieving both reactivity and controllability in trajectory generation. Following this work, Zhou et al.~\cite{zhou2025decoupled} proposed Nexus, a decoupled diffusion framework that models structured and independent noise states together with a noise-aware scheduling mechanism. By enabling timely environmental updates during denoising, Nexus allows agents to adapt their behaviors interactively within evolving traffic contexts, leading to more stable and reactive closed-loop simulation.

Inspired by these developments, we present nuPlan-R, a reactive closed-loop planning benchmark designed to enable more realistic and fair evaluation of autonomous driving planners. Our learning-based reactive agents are seamlessly integrated into the nuPlan simulation framework as configurable components, replacing rule-based IDM agents with diffusion-based reactive models following the Nexus architecture~\cite{zhou2025decoupled}. To further improve simulation efficiency, we introduce an interaction-aware agent selection mechanism that focuses on agents most relevant to the ego vehicle. Moreover, we extend the benchmark with two complementary metrics, Success Rate (SR) and All-Core Pass Rate (PR), to assess a planner’s robustness and overall performance balance across multiple dimensions. Extensive experiments demonstrate that our reactive agents exhibit superior realism and behavioral diversity compared to IDM-based agents. Furthermore, we reimplement a collection of rule-based, learning-based, and hybrid planning approaches within the nuPlan-R benchmark, revealing that nuPlan-R more effectively highlights the interactive advantages of learning-based planners in complex scenarios. These results confirm that nuPlan-R provides a more reliable and insightful benchmark for closed-loop planning evaluation.

In summary, the main contributions of this paper include: 
\begin{itemize}
    \item We propose nuPlan-R, a new reactive closed-loop planning benchmark that seamlessly integrates learning-based reactive agents and an interaction-aware agent selection mechanism into the nuPlan simulation framework as configurable components, enabling more realistic and human-like multi-agent interactions.
    \item We extend the benchmark with two additional metrics, Success Rate and All-Core Pass Rate, to more effectively evaluate the robustness of planners and balance across various dimensions of driving performance.
    \item We reimplement and comprehensively evaluate a diverse set of planning approaches in both the original nuPlan and the proposed nuPlan-R benchmarks, demonstrating that nuPlan-R provides a more reliable assessment of planners’ interaction capabilities in complex traffic environments.
\end{itemize}

\section{RELATED WORK}
\label{sec:related work}

\subsection{Planning Benchmark for Autonomous Vehicles} 
Table~\ref{tab:comparison} shows a comparison between our benchmark and other autonomous driving benchmarks. Early benchmarks for autonomous vehicle motion planning, such as Waymo~\cite{ettinger2021large} and nuScenes~\cite{caesar2020nuscenes}, primarily focus on short-term motion forecasting under open-loop evaluation, where displacement-based metrics are used to assess prediction accuracy. However, these open-loop evaluations neglect the closed-loop interactions between the ego vehicle and its surroundings, and therefore fail to evaluate the long-term safety and feasibility of the generated trajectories.
To address these limitations, closed-loop planning benchmarks such as nuPlan~\cite{caesar2021nuplan} and Waymax~\cite{gulino2023waymax} introduce scenario-based planning metrics and reactive multi-agent simulation to enable more realistic evaluation of long-horizon planning performance. However, their reactive agents are typically rule-based, for example, following the IDM~\cite{treiber2000congested}, a classical car-following model that regulates acceleration based on desired speed and time headway. Although computationally efficient, IDM agents lack diversity and human-like decision-making, often leading to overly simplified and less interactive traffic behaviors.
Several End-to-End (E2E) driving benchmarks, such as CARLA~\cite{dosovitskiy2017carla} and NAVSIM~\cite{dauner2024navsim}, also adopt closed-loop evaluation. However, they either rely on non-reactive agents~\cite{dauner2024navsim, jia2024bench2drive} or use IDM-based reactive models \cite{dosovitskiy2017carla, you2024bench2drive}, and thus inherit similar limitations. Overall, current benchmarks, whether behavior-focused planning benchmarks or perception-to-control E2E driving benchmarks, suffer from a common limitation of failing to model realistic multi-agent interactions. 
In contrast, our proposed nuPlan-R benchmark incorporates learning-based reactive agents that produce diverse, human-like behaviors in closed-loop multi-agent simulation, providing a more realistic and fair environment for evaluating autonomous driving planners.

\subsection{Multi-Agent Traffic Simulation}
Early data-driven multi-agent simulation learn from real driving logs to replace rule-based heuristics and better capture interactive behaviors \cite{suo2021trafficsim, bergamini2021simnet, xu2022bits}. 
Diffusion-based approaches further improve realism and controllability \cite{peng2025diffusion}, allowing multi-agent trajectory generation conditioned on Signal Temporal Logic (STL) \cite{zhong2023guided}, hand-crafted cost functions \cite{jiang2023motiondiffuser, peng2025safety}, Reinforcement Learning (RL) rewards \cite{xie2024advdiffuser}, or even LLM-based guidance \cite{zhong2023language, peng2025ld}. However, these models still struggle with long-term closed-loop stability due to covariate shift between the training data distribution and the distribution of trajectories generated by the model during rollouts, where small prediction errors can accumulate over time and lead to severe performance degradation.
In contrast, Next-Token Prediction (NTP) models formulate multi-agent motion as discrete sequence modeling \cite{philion2023trajeglish, seff2023motionlm, wu2024smart, zhang2025closed}, demonstrating stronger long-horizon closed-loop simulation performance, but offering less direct controllability.
Recently, Diffusion Forcing \cite{chen2024diffusion} unifies NTP models and full-sequence diffusion under a partial-masking view. By training with independent per-token noise levels in a causal architecture, it combines the reactive generation of autoregressive models with the controllability of diffusion models.
Inspired by this perspective, Zhou et al.\ proposed Nexus \cite{zhou2025decoupled}, a noise-decoupled diffusion framework that introduces independent yet structured noise states and a noise-aware scheduling mechanism, enabling dynamic scene updates and real-time adaptation to environmental changes for more stable and realistic closed-loop multi-agent traffic simulation.
Building upon these insights, we integrate noise-decoupled diffusion-based reactive agents into the nuPlan framework, replacing conventional rule-based IDM agents and enabling a new form of closed-loop planning benchmark through reactive multi-agent simulation.

\begin{table}[t]
\footnotesize
\centering
\renewcommand{\arraystretch}{1.2}
\caption{Comparison of autonomous driving benchmarks. 
\textcolor{green}{\ding{51}} / \textcolor{red}{\ding{55}} indicate supported / unsupported. 
In the Reactive column, 
\textcolor{orange}{\ding{51}} represents rule-based reactive agents, 
and \textcolor{green}{\ding{51}} represents learning-based reactive agents.}
\begin{tabular}{c|c|c|c|c}
\Xhline{1.0pt}
Benchmark & \makecell{Real-world \\ Data} & \makecell{Closed-\\Loop} & Reactive & Task \\
\Xhline{1.0pt}
Waymo           & \cmark & \xmark & \xmark & Planning \\
nuScenes        & \cmark & \xmark & \xmark & Planning \\
nuPlan          & \cmark & \cmark & \cmid & Planning \\
Waymax          & \cmark & \cmark & \cmid & Planning \\
NAVSIM          & \cmark & \cmark & \xmark & E2E Driving \\
CARLA           & \xmark & \cmark & \cmid & E2E Driving \\
Bench2Drive     & \xmark & \cmark & \xmark & E2E Driving \\
Bench2Drive-R   & \cmark & \cmark & \cmid & E2E Driving \\
nuPlan-R (ours) & \cmark & \cmark & \cmark & Planning \\
\Xhline{1.0pt}
\end{tabular}
\label{tab:comparison}
\end{table}

\begin{figure*}[!t]
        \centering
        \includegraphics[trim={72 240 80 70}, clip, width=\linewidth]{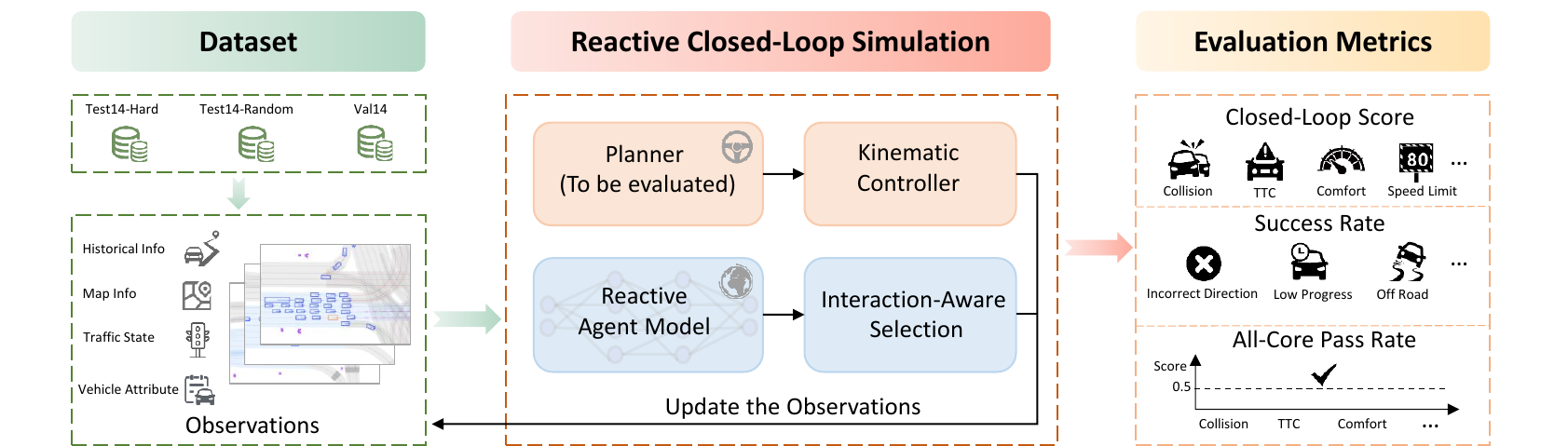}
        % \vspace{-1.0em}
        \caption{A road map of reactive closed-loop planning benchmark: nuPlan-R. The Reactive Agent Model represents our trained learning-based reactive agent, which replaces the traditional rule-based IDM agents to enable realistic and interactive traffic simulation.}
        % \vspace{-0.5em}
\label{fig:road_map}
\end{figure*}

\section{METHODOLOGY}
\label{sec:methodology}
The road map of our nuPlan-R benchmark, as shown in Fig.~\ref{fig:road_map}. It replaces the rule-based IDM reactive agents with our learning-based reactive models and extends the nuPlan benchmark with additional metrics for more realistic and comprehensive closed-loop planning evaluation.
\subsection{Task Description}
This work aims to develop a reactive multi-agent simulation framework for the realistic and reliable evaluation of autonomous planners. Specifically, each agent processes a stream of historical observations as inputs for trajectory generation. The simulation then updates each agent's observations at each time step based on the newly predicted multi-agent trajectories, thereby enabling reactive interactions among all agents.

\textbf{Problem Formulation.} We simulate a traffic scenario involving $N$ agents, where one agent represents the ego vehicle, controlled by the planner $\pi$, and the future trajectories of the remaining $N-1$ agents are predicted by our model. The state of each agent at time $t$ is represented as
$
s^{i}_t = \left(x^{i}_t, y^{i}_t, \sin(\theta^{i}_t), \cos(\theta^{i}_t), v^{x}_t, v^{y}_t, l^{i}, w^{i}\right).
$
The past trajectories of all agents over the past $T_\mathit{hist}$ time steps are denoted as 
$
\textit{\textbf{x}} = \{s_{t-T_\mathit{hist}}, s_{t-T_\mathit{hist}+1}, ..., s_{t}\}.
$
Additionally, we define the map information as $\textit{\textbf{m}}$.

At each simulation step, the ego vehicle state is updated by $s^{0}_{t+1} = \mathcal{C}(\pi(\textit{\textbf{m}},\textit{\textbf{x}}))$, where $\mathcal{C}$ represents the controller that applies control actions based on the planner $\pi$. The surrounding vehicles update their states using the reactive agent model $\{s^{i}_{t+1}\}^{N-1}_{i=1} = \mathcal{F}(\textit{\textbf{m}},\textit{\textbf{x}})$. The updated states $s^{i}_{t+1}$ constitute the new observations at time $t+1$.

In this closed-loop setup, both the ego planner and surrounding agents update their state at 10 Hz, closely mimicking the dynamic and interactive behaviors in real-world traffic testing.

\begin{table*}[!t]
\centering
\renewcommand{\arraystretch}{1.3}
\caption{Evaluation of multi-agent traffic simulation performance on the Test14-Hard dataset. Our reactive agent achieves superior distribution realism, diversity, and behavior plausibility compared to rule-based (IDM) and diffusion-based baselines.}
\begin{tabularx}{0.95\textwidth}{
    >{\centering\arraybackslash}X
    |>{\centering\arraybackslash}X
    |>{\centering\arraybackslash}X
    |>{\centering\arraybackslash}X
    |>{\centering\arraybackslash}X
    |>{\centering\arraybackslash}X
    |>{\centering\arraybackslash}X
}
\Xhline{1.0pt} 
\multirow{2}{*}{\textbf{\makecell{\\Method}}} 
& \multicolumn{2}{c|}{\textbf{Distribution Realism}} 
& \textbf{Diversity}
& \multicolumn{3}{c}{\textbf{Behavior Plausibility}} \\ 
\cline{2-7}
\rule{0pt}{4ex}
& \makecell{TTC JSD \\ ($\times 10^{-2}$) $\downarrow$} & \makecell{Cluster-Traj. \\ FID $\downarrow$} & \makecell{Shannon \\ Entropy $\uparrow$} & \makecell{ Off-Road \\ (\%) $\downarrow$} & \makecell{Other-Other Coll \\ (\%) $\downarrow$} & \makecell{ Ego-Other Coll \\ (\%) $\downarrow$} \\
\Xhline{1.0pt}
Log-Replay          & --    & --    & 0.988 & 4.36  & 0.40 & 1.53 \\
IDM \cite{helbing1998generalized}                 & 6.15  & 2.52  & 0.841 & \textbf{0.0}   & \textbf{0.17} & 2.33 \\
FDM-N \cite{zhengdiffusion}   & 6.20  & 3.79  & 0.979 & 10.18 & 0.68 & 8.92 \\
Nexus \cite{zhou2025decoupled}              & 1.80  & 2.17  & 0.975 & 5.53  & 0.42 & 1.75 \\
Ours                & \textbf{1.49}  & \textbf{1.84}  & \textbf{0.987} & 4.79  & 0.44 & \textbf{1.45} \\
\Xhline{1.0pt} 
\end{tabularx}
\label{tab:multi_gen_quality}
\end{table*}

\begin{figure*}[!t]
        \centering
        \includegraphics[trim={1.8cm 6.5cm 2cm 6cm}, clip, width=\linewidth]{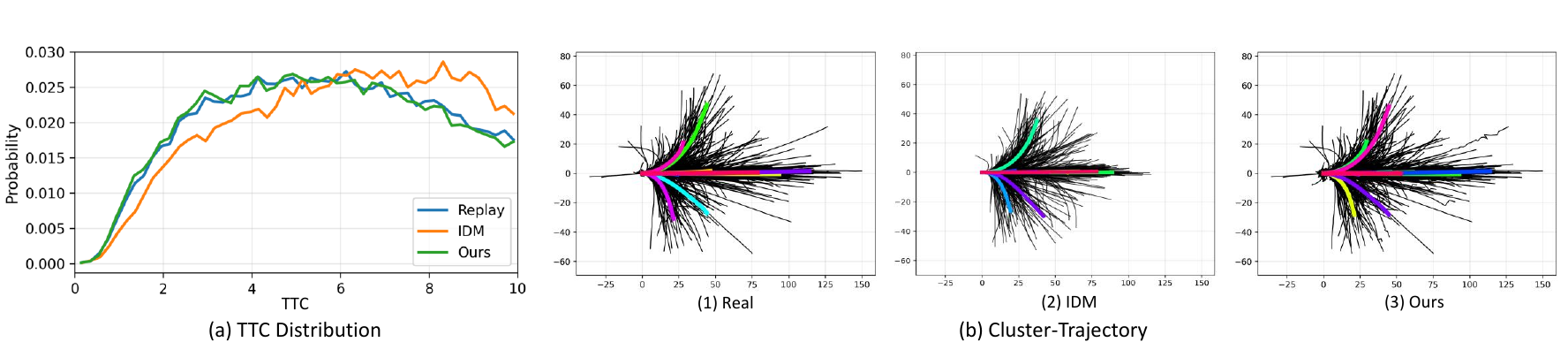}
        % \vspace{-1.0em}
        \caption{Visual illustration of the results. 
(a) TTC Distribution: Comparison of TTC distributions between log-replay, rule-based IDM, and our learning-based reactive agent, showing that our model better aligns with real-world statistics. 
(b) Cluster-Trajectory: Visualization of representative trajectory clusters for (1) log-replay data, (2) IDM simulation, and (3) our reactive simulation, demonstrating that our model captures richer and more realistic multi-agent motion patterns.}

        % \vspace{-0.5em}
\label{fig:trajectory_quality}
\end{figure*}

\subsection{Reactive Multi-Agent Model}  
\textbf{Architecture.} We train a noise-decoupled diffusion-based reactive agent model following the Nexus framework~\cite{zhou2025decoupled}. The model utilizes a Diffusion Transformer (DiT) backbone~\cite{peebles2023scalable}, which processes a vectorized scene representation where agent tokens and map features are encoded and processed through cross-attention mechanisms.

To enable reactivity, we also assign an independent noise level \( k_{a,t} \in (0,1] \) to each agent-time token. Given a clean data input \( x^0 \), noise is added to each token as follows:
\begin{equation}
x^t = g(x^0, k),
\end{equation}
where \( g \) represents the forward process of diffusion models that perturbs the input with noise \cite{ho2020denoising}. The model is trained to denoise the perturbed input by minimizing the noise prediction loss:
\begin{equation}
\min_{\theta} \mathbb{E}\left[\|\epsilon - \epsilon_{\theta}(g(x^0,k), c, k)\|_2^2\right],
\end{equation}
where \( c \) represents the vectorized map tensor, which serves as the conditioning input.

During inference, lower noise levels are assigned to historical and goal tokens, while higher noise levels are applied to future tokens, thereby facilitating flexible trajectory generation. A noise-aware scheduling matrix \( K \) controls the adaptive updates of the scene chunks: after each denoising step, low-noise tokens are replaced with new high-noise frames~\cite{zhou2025decoupled}. Any environmental changes to the agent tensor can directly replace the agent state and reduce the noise, enabling the scene to dynamically adapt during the simulation.

\textbf{Interaction-Aware Selection Mechanism.} To improve computational efficiency and mitigate covariate shift between generated and real-world traffic distributions, we design an interaction-aware agent selection algorithm. We compute an interaction intensity score between each surrounding agent and the ego vehicle by jointly considering relative distance, heading difference, and velocity difference. Specifically, the interaction score $\mathcal{I}$ for each agent $i$ is computed as:
\begin{equation}
\footnotesize
\mathcal{I}_i = w_{d} \! \cdot \! \exp\!\left(\!-\frac{d_i}{d_{\text{thresh}}}\!\right)\!+\!w_{v}\!\cdot\! \frac{|v_{i,\text{rel}}|}{\max(|v_{\text{rel}}|)}\!+\!w_{h}\!\cdot\!\left(\!1\!-\!|\cos(\Delta \theta_i)| \right),
\end{equation}
where $d_i$ is the distance between the ego vehicle and agent $i$, $v_{i,\text{rel}}$ is the relative velocity of agent $i$ with respect to the ego vehicle, $\Delta \theta_i$ is the difference in heading between the ego vehicle and agent $i$, $d_{\text{thresh}}$ is a distance threshold for interaction relevance, and $w_{d}, w_{v}, w_{h}$ are weights for distance, velocity, and heading, respectively. After computing the interaction intensity for all agents, the top-$k$ agents with the highest interaction scores are selected as reactive agents. These agents are updated through our diffusion-based model rollout, while the remaining agents follow log-replay updates from log-recorded trajectories. To further improve the quality of the reactive agents' trajectories, a post-smoothing optimization process is applied to the selected agents, ensuring smoother and more physically plausible motion while maintaining their interactive behaviors. This hybrid rollout strategy ensures that realistic local interactions around the ego vehicle are preserved, while maintaining stable large-scale traffic flow with minimal computational overhead.

\textbf{Integration.} 
Our learning-based reactive agents are seamlessly integrated into the nuPlan simulation framework as configurable components. 
Users can easily switch the simulation mode by modifying the reactive agent configuration, replacing the rule-based IDM with our learning-based reactive agents. 
During closed-loop evaluation, both the ego planner and reactive agents update their trajectories synchronously at 10\,Hz, forming a unified, data-driven environment that simulates realistic multi-agent interactions. 

\subsection{Benchmark Metrics}
\label{sec:benchmark_metrics}
The original nuPlan benchmark primarily adopts a Closed-Loop Score (CLS), which aggregates multiple predefined metrics, such as comfort, collision, time-to-collision (TTC), and progress efficiency, to evaluate overall planning performance.
We extend the benchmark metrics of \textit{nuPlan-R} by introducing two additional indicators, namely SR and PR, thus expanding the benchmark to three complementary metrics in total.

\textbf{Success Rate.}
This metric measures a planner’s robustness in maintaining safe and feasible trajectories throughout a closed-loop rollout. It is computed as the proportion of scenarios where the CLS is non-zero, indicating that the ego vehicle avoids major failures such as severe collisions, off-road departures, wrong-way driving, or extremely low progress. A higher SR reflects a planner’s reliability in maintaining feasible and safe behaviors across diverse scenarios. Similar formulations have also been adopted by recent planning approaches such as CALMM \cite{yao2024calmm}, underscoring its effectiveness as an indicator of planning robustness.

\textbf{All-Core Pass Rate.}
To further evaluate whether a planner performs consistently across all aspects rather than excelling in only a few, we introduce the All-Core Pass Rate. It measures the proportion of scenarios where all core sub-metrics (such as safety, comfort, and efficiency) achieve normalized scores above 0.5. This indicator highlights the planner’s overall balance and compliance in diverse driving scenarios.

\begin{figure*}[!t]
        \centering
        \includegraphics[trim={20 172 50 0}, clip, width=\linewidth]{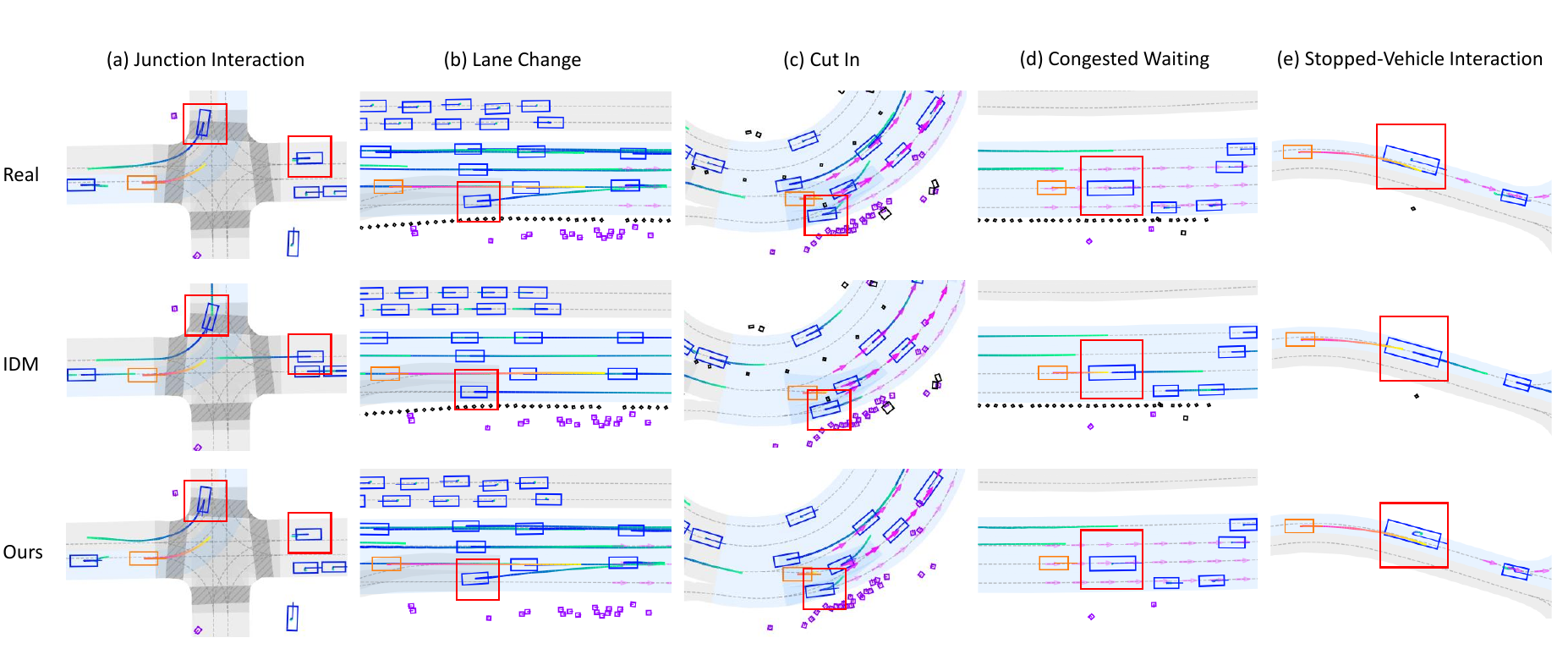}
        % \vspace{-1.0em}
\caption{Qualitative comparison of reactive closed-loop simulation behaviors between rule-based IDM agents and our learning-based agents across five representative scenarios. Our model produces more realistic and reactive interactions that better resemble real-world driving behaviors.}
        % \vspace{-0.5em}
\label{fig:scenarios}
\end{figure*}

\section{EXPERIMENTAL RESULTS}
\label{sec:experiment}
We conduct two sets of experiments to validate our benchmark.
First, we evaluate the multi-agent simulation behavior to verify the realism, diversity, and plausibility of our reactive agents compared with representative baselines.
Second, we assess planner performance under both the original nuPlan and our proposed nuPlan-R benchmarks to examine how realistic reactive interactions influence closed-loop planning evaluation.
\subsection{Evaluation of Multi-Agent Simulation Behavior}
\textbf{Evaluation Metrics.}
We assess the quality of generated multi-agent behaviors from three aspects: Distribution Realism, Diversity, and Behavior Plausibility.
\begin{itemize}
    \item Distribution Realism: Measures the similarity between generated and real-world driving behaviors. TTC Jensen–Shannon Divergence (JSD) \cite{lin2002divergence} quantifies distributional gaps in TTC histograms, while Cluster-Trajectory Fréchet Inception Distance (FID) \cite{heusel2017gans} evaluates the similarity between trajectory clusters grouped by geometric and kinematic attributes. Lower values indicate higher realism.

    \item Diversity: Shannon Entropy \cite{shannon1948mathematical} computed from the similarity matrix of all trajectory pairs, characterized by endpoints, path length, and curvature, quantifies behavioral richness; higher entropy denotes greater diversity in motion patterns.

    \item Behavior Plausibility: Off-Road Rate, measuring the proportion of vehicles leaving drivable areas; Other-Other Collision Rate, capturing collisions among surrounding vehicles; and Ego-Other Collision Rate, indicating collisions involving the ego vehicle. Lower values imply more physically consistent and plausible behaviors.
\end{itemize}

\textbf{Experimental Setup.}
We compare our models with three representative baseline models and present the results of the Log-Replay experiment as a reference. The baselines include IDM \cite{helbing1998generalized}, a rule-based reactive algorithm; FDM-N \cite{zhengdiffusion}, a full-sequence diffusion-based model that operates without frame-by-frame updates (non-reactive); and Nexus, a diffusion-based reactive simulation. These methods were evaluated through an 8-second closed-loop simulation of multiple surrounding vehicles. In these closed-loop experiments, we uniformly used the diffusion planner \cite{zhengdiffusion} to interact with the surrounding vehicles.

\begin{table*}[!t]
\footnotesize
\centering
\renewcommand{\arraystretch}{1.35}
\caption{Quantitative evaluation of reactive closed-loop simulation performance of different planning approaches on the original nuPlan and our proposed nuPlan-R benchmarks using the Test14-Random and Test14-Hard datasets. The evaluation includes the standard CLS as well as the two additional metrics introduced in nuPlan-R: SR and PR.}
\begin{tabular}{c|c|c|c|c|c|c|c|c|c}
\Xhline{1.0pt}
\multirow{3}{*}{\makecell{\\ \textbf{Type}}} & \multirow{3}{*}{\makecell{\\ \textbf{Planner}}}
& \multicolumn{4}{c|}{\textbf{Test14-Random}}
& \multicolumn{4}{c}{\textbf{Test14-Hard}} \\ \cline{3-10}
& & \multicolumn{1}{c|}{\textbf{nuPlan}} & \multicolumn{3}{c|}{\textbf{nuPlan-R}}
  & \multicolumn{1}{c|}{\textbf{nuPlan}} & \multicolumn{3}{c}{\textbf{nuPlan-R}} \\ 
\cline{3-10}
\rule{0pt}{4ex}
& & \makecell{IDM-\\R-CLS $\uparrow$}
  & \makecell{Diffusion- \\R-CLS $\uparrow$}
  & \makecell{Diffusion- \\R-SR(\%) $\uparrow$}
  & \makecell{Diffusion-\\R-PR(\%) $\uparrow$}
  & \makecell{IDM-\\R-CLS $\uparrow$}
  & \makecell{Diffusion-\\R-CLS $\uparrow$}
  & \makecell{Diffusion-\\R-SR(\%) $\uparrow$}
  & \makecell{Diffusion-\\R-PR(\%) $\uparrow$} \\
\Xhline{0.8pt}
\multirow{1}{*}{Expert}
& Log-Replay          & 75.86 & 94.25 & 97.29 & 90.73 & 68.80 & 85.85 & 91.54 & 78.67 \\
\Xhline{0.8pt}
\multirow{2}{*}{\makecell{Rule-\\Based}}
& IDM \cite{helbing1998generalized}    & 72.42 & 70.75 & 76.63 & 61.68 & 62.26 & 57.49 & 67.66 & 43.49 \\
& PDM-Closed \cite{dauner2023parting}  & 91.64 & 90.62 & 95.40 & 82.76 & 75.19 & 67.33 & 80.88 & 46.69 \\
\Xhline{0.8pt}
\multirow{5}{*}{\makecell{Learning-\\Based}}
& GC-PGP \cite{hallgarten2023prediction}   & 56.46 & 59.79 & 69.35 & 44.83 & 42.80 & 46.13 & 55.51 & 29.78 \\
& UrbanDriverOL \cite{scheel2022urban}     & 60.92 & 66.19 & 74.33 & 50.57 & 49.06 & 55.71 & 65.07 & 40.81 \\
& RasterModel  \cite{caesar2021nuplan}     & 68.64 & 66.28 & 72.41 & 52.87 & 52.44 & 51.12 & 58.09 & 36.40 \\
& PlanTF   \cite{cheng2024rethinking}      & 78.86 & 84.98 & 89.65 & 78.16 & 60.34 & 73.41 & 79.41 & 67.28 \\
& Diffusion-Planner \cite{zhengdiffusion}  & 83.21 & 88.46 & 93.63 & 82.47 & 68.83 & 75.70 & 84.19 & 64.70 \\
\Xhline{0.8pt}
Hybrid % \multirow{4}{*}{Hybrid} 
& PDM-Hybrid  \cite{dauner2023parting}     & 91.29 & 90.48 & 95.40 & 82.76 & 76.07 & 67.44 & 80.15 & 47.06 \\
% & GameFormer          & 77.80 & --    & --    & --    & 61.53 & --    & --    & --    \\
% & Diffusion-ES        & 87.18 & --    & --    & --    & 77.75 & --    & --    & --    \\
% & PLUTO               & 89.94 & --    & --    & --    & 75.75 & --    & --    & --    \\
% \Xhline{0.8pt}
% \makecell{LLM-\\based}
% & CALMM-Drive         & 87.11 & --    & --    & --    & 78.13 & --    & --    & --    \\
\Xhline{1.0pt}
\end{tabular}
\label{tab:planner_results}
\end{table*}

\textbf{Results.}
The quantitative performance of different methods on multi-agent trajectory generation is presented in Table~\ref{tab:multi_gen_quality}. TTC JSD and Cluster-Trajectory FID metrics demonstrate that our model achieves the highest level of realism, exhibiting human-like driving behaviors. Regarding Diversity, the entropy of trajectory geometry shows that our method produces the richest variety of motion patterns among all baseline models, significantly surpassing IDM’s overly rigid and rule-driven behavior. Although IDM achieves the lowest Off-Road and Collision rates, which indicates strong compliance with driving constraints, this level of safety is obtained at the cost of realism and reactivity. Because IDM relies solely on longitudinal interactions with the preceding vehicle, it fails to capture the multi-directional dynamic responses between agents and the ego vehicle. Consequently, IDM exhibits delayed reactions in complex traffic scenarios, as reflected by its higher Ego-Other Collision rate. Similarly, the diffusion-based non-reactive baseline lacks frame-wise action updates, leading to poor responsiveness and increased risk of active collisions. Notably, our model slightly outperforms Nexus, suggesting that the proposed interaction-aware agent selection mechanism and trajectory post-smoothing optimization further improve stability and behavior plausibility in closed-loop simulation. Overall, our diffusion-based closed-loop reactive model achieves a better balance among realism, diversity, and safety, enabling more realistic and reliable evaluation for planning algorithms.

Moreover, as shown in Fig.~\ref{fig:trajectory_quality}, the TTC distribution visualization in Fig.~\ref{fig:trajectory_quality}(a) further confirms the advantage of our model. IDM displays an overly conservative pattern with higher TTC values, which deviates from real-world interaction frequencies, while our TTC curve more closely aligns with the real distribution. Fig.~\ref{fig:trajectory_quality}(b) visualizes the clustered trajectory modes, showing that IDM trajectories are concentrated in fewer motion patterns. In contrast, our model recovers a diverse and realistic set of driving behaviors that are consistent with real-world clusters. 

Fig.~\ref{fig:scenarios} qualitatively compares the reactive closed-loop simulation behaviors of IDM agents and our learning-based agents across five representative scenarios. The orange vehicle denotes the ego vehicle, and the red boxes highlight the main interacting vehicles. In the complex junction scenario shown in (a), the IDM agent strictly follows the lane centerline, resulting in active overlap into the ego’s path and a potential collision risk. In contrast, our reactive model exhibits a ``nudge" behavior, adjusting its turning path to avoid conflicts and more closely mimic natural human driving. Moreover, in the IDM simulation, the rightward vehicle always forces the right-of-way without perceiving the ego car as an interactive agent, resulting in rigid and non-cooperative motion, whereas our model simulates realistic yielding behavior consistent with human driving at this junction. In (b) and (c), our model captures lateral interactions such as lane changes and cut-ins, whereas the IDM agent only produces monotonic, lane-centered trajectories with limited responsiveness. In (d) and (e), our model effectively simulates scenarios involving stationary vehicles, such as queuing in congestion and unexpected stopped obstacles, allowing the planner’s decision-making and reactivity to be evaluated under realistic waiting and emergency conditions. The IDM agent, limited by its purely longitudinal dynamics, cannot accurately model these behaviors. Collectively, these visualizations highlight the enhanced realism and interactivity of our learning-based reactive agent.

These qualitative and quantitative observations collectively demonstrate that our reactive agents model enables more human-like multi-agent interactions and generates more realistic behaviors in closed-loop simulation.

\subsection{Results of Planner Models in nuPlan-R Benchmark}
\textbf{Experimental Setup.}
We evaluate a diverse set of planners on both the nuPlan benchmark and our proposed nuPlan-R benchmark under a reactive closed-loop setting, using the Test14-Random and Test14-Hard datasets. These datasets cover 14 scenario types from the nuPlan Planning Challenge, where Test14-Hard particularly focuses on long-tail driving scenarios. In the original nuPlan benchmark, surrounding vehicles are controlled by a rule-based reactive agent based on the IDM, while in our nuPlan-R benchmark, they are driven by the proposed learning-based reactive agent, which enables more realistic interactions. In addition to the standard CLS metric, we also report two additional metrics, namely SR and PR metrics, as described in Section~\ref{sec:benchmark_metrics}. In all closed-loop simulations, we further employ a Linear Quadratic Regulator (LQR) controller for trajectory tracking, and the resulting control outputs are applied to a kinematic bicycle model to propagate the vehicle dynamics.

\textbf{Planner Models.}
We evaluate a varied collection of planners, including rule-based, learning-based, and hybrid approaches, to assess their performance both in the nuPlan benchmark and in our proposed nuPlan-R benchmark.
\begin{itemize}
\item IDM~\cite{helbing1998generalized}: A classical car-following model where the vehicle dynamically adjusts its acceleration based on the relative distance and speed difference with the leading vehicle.
\item PDM-Closed~\cite{dauner2023parting}: A rule-based planner that integrates IDM with centerline-based trajectory proposals evaluated through a simulate and score loop to select the optimal closed-loop plan.
\item GC-PGP~\cite{hallgarten2023prediction}: A goal-conditioned planning method that extends the graph-based learning-based planner by incorporating goal information to ensure route adherence while preserving multi-modal motion diversity.
\item UrbanDriver~\cite{scheel2022urban}: A vectorized imitation learning planner that employs PointNet-based polyline encoders to process vectorized map and agent features. Here, we use its open-loop re-implementation in nuPlan.
\item RasterModel~\cite{caesar2021nuplan}: A convolutional neural network (CNN)-based planner that utilizes rasterized bird's-eye-view (BEV) images of the driving scenarios to generate motion plans.
\item PlanTF~\cite{cheng2024rethinking}: A Transformer-based imitation learning planner that encodes agents, HD-map features, and ego state through attention mechanisms, achieving strong closed-loop performance without rule-based post-processing.
\item Diffusion-Planner~\cite{zhengdiffusion}: A diffusion-based planner built on a DiT architecture that jointly generates ego trajectories and multi-agent predictions, achieving strong performance without rule-based post-processing.
\item PDM-Hybrid~\cite{dauner2023parting}: A hybrid planner that integrates PDM-Closed with learned ego-forecasting adjustments, achieving high accuracy by dynamically adjusting predictions based on surrounding vehicle interactions.
\end{itemize}

\textbf{Results.}
As shown in Table~\ref{tab:planner_results}, a clear distinction can be observed between the performance trends of rule-based and learning-based planners across the nuPlan and our proposed nuPlan-R benchmarks. The rule-based planners exhibit a drop in CLS when evaluated in nuPlan-R, and the hybrid planner, which integrates rule-based post-processing to boost its performance in the original nuPlan benchmark, also experiences a decrease in CLS within nuPlan-R. These findings indicate that while rule-based algorithms may take advantage of the simplified reactive agent dynamics in the original benchmark, such benefits do not generalize to the more realistic and interactive closed-loop simulation of nuPlan-R. In contrast, learning-based planners show clear performance improvements in the nuPlan-R benchmark. The performance gap becomes even more pronounced on the Test14-Hard dataset, demonstrating that nuPlan-R provides a fairer and more effective evaluation of planners' decision-making and interaction capabilities in complex traffic scenarios.

Beyond CLS, the additional SR and PR metrics provide complementary insights into planner robustness and overall balance. SR reveals the fragility that may be obscured by CLS, as it is more sensitive to whether a planner can reliably avoid catastrophic failures. PR, on the other hand, evaluates whether a planner performs consistently across all dimensions, rather than achieving high overall scores by emphasizing only a few aspects. On the Test14-Hard dataset, a more significant reordering of planner rankings appears in the SR and PR metrics. For instance, PDM-Hybrid maintains a competitive CLS, but its PR drops considerably, reflecting unbalanced behavior. Together, these three metrics provide a more comprehensive reflection of planners’ true performance under complex interactive scenarios.

\section{CONCLUSIONS}
\label{sec:conclusion}
In this paper, we presented nuPlan-R, a new reactive closed-loop planning benchmark that integrates learning-based reactive agents into the nuPlan framework for realistic and fair evaluation of autonomous planners. Our noise-decoupled diffusion agents enable diverse and human-like multi-agent behaviors while maintaining stable closed-loop simulation. We also introduced two complementary metrics to provide a more comprehensive assessment of planner robustness and balance. Extensive experiments demonstrate that nuPlan-R produces more realistic simulations and better reveals planner performance differences in complex interactive scenarios.
In future work, we plan to incorporate large language or vision language models to further enhance the diversity, controllability, and interpretability of reactive multi-agent traffic simulation.
% \section*{ACKNOWLEDGMENT}

%%%%%%%%%%%%%%%%%%%%%%%%%%%%%%%%%%%%%%%%%%%%%%%%%%%%%%%%%%%%%%%%%%%%%%%%%%%%%%%%
\bibliographystyle{ieeetr}
\bibliography{reference}

@string{ICRA = {the Proceedings of IEEE International Conference on Robotics and Automation}}

@string{CVPR = {the Proceedings of the IEEE/CVF Conference on Computer Vision and Pattern Recognition}}

@string{ICCV = {the Proceedings of the IEEE/CVF International Conference on Computer Vision}}

@string{IROS = {the Proceedings of IEEE/RSJ International Conference on Intelligent Robots and Systems}}

@string{CORL = {the Proceedings of Conference on Robot Learning}}

@string{ICLR = {the Proceedings of International Conference on Learning Representations}}

@string{NIPS = {the Proceedings of Advances in Neural Information Processing Systems}}

@inproceedings{ettinger2021large,
  title={Large scale interactive motion forecasting for autonomous driving: The waymo open motion dataset},
  author={Ettinger, Scott and Cheng, Shuyang and Caine, Benjamin and Liu, Chenxi and Zhao, Hang and Pradhan, Sabeek and Chai, Yuning and Sapp, Ben and Qi, Charles R and Zhou, Yin and others},
  booktitle=ICCV,
  pages={9710--9719},
  year={2021}
}

@inproceedings{caesar2020nuscenes,
  title={{nuScenes}: A multimodal dataset for autonomous driving},
  author={Caesar, Holger and Bankiti, Varun and Lang, Alex H and Vora, Sourabh and Liong, Venice Erin and Xu, Qiang and Krishnan, Anush and Pan, Yu and Baldan, Giancarlo and Beijbom, Oscar},
  booktitle=CVPR,
  pages={11621--11631},
  year={2020}
}

@article{caesar2021nuplan,
  title={{nuPlan}: A closed-loop ml-based planning benchmark for autonomous vehicles},
  author={Caesar, Holger and Kabzan, Juraj and Tan, Kok Seang and Fong, Whye Kit and Wolff, Eric and Lang, Alex and Fletcher, Luke and Beijbom, Oscar and Omari, Sammy},
  journal={arXiv preprint arXiv:2106.11810},
  year={2021}
}

@inproceedings{gulino2023waymax,
  title={{Waymax}: An accelerated, data-driven simulator for large-scale autonomous driving research},
  author={Gulino, Cole and Fu, Justin and Luo, Wenjie and Tucker, George and Bronstein, Eli and Lu, Yiren and Harb, Jean and Pan, Xinlei and Wang, Yan and Chen, Xiangyu and others},
  booktitle=NIPS,
  volume={36},
  pages={7730--7742},
  year={2023}
}

@article{treiber2000congested,
  title = {Congested traffic states in empirical observations and microscopic simulations},
  author = {Treiber, Martin and Hennecke, Ansgar and Helbing, Dirk},
  journal = {Physical Review E},
  volume = {62},
  issue = {2},
  pages = {1805--1824},
  numpages = {0},
  year = {2000}
}

@inproceedings{dosovitskiy2017carla,
  title={{CARLA}: An open urban driving simulator},
  author={Dosovitskiy, Alexey and Ros, German and Codevilla, Felipe and Lopez, Antonio and Koltun, Vladlen},
  booktitle={Conference on Robot Learning},
  pages={1--16},
  year={2017},
}

@inproceedings{dauner2024navsim,
  title={{NAVSIM}: Data-driven non-reactive autonomous vehicle simulation and benchmarking},
  author={Dauner, Daniel and Hallgarten, Marcel and Li, Tianyu and Weng, Xinshuo and Huang, Zhiyu and Yang, Zetong and Li, Hongyang and Gilitschenski, Igor and Ivanovic, Boris and Pavone, Marco and others},
  booktitle=NIPS,
  volume={37},
  pages={28706--28719},
  year={2024}
}

@inproceedings{jia2024bench2drive,
  title={{Bench2Drive}: Towards multi-ability benchmarking of closed-loop end-to-end autonomous driving},
  author={Jia, Xiaosong and Yang, Zhenjie and Li, Qifeng and Zhang, Zhiyuan and Yan, Junchi},
  booktitle=NIPS,
  volume={37},
  pages={819--844},
  year={2024}
}

@article{you2024bench2drive,
  title={{Bench2Drive-R}: Turning real world data into reactive closed-loop autonomous driving benchmark by generative model},
  author={You, Junqi and Jia, Xiaosong and Zhang, Zhiyuan and Zhu, Yutao and Yan, Junchi},
  journal={arXiv preprint arXiv:2412.09647},
  year={2024}
}

@inproceedings{suo2021trafficsim,
  title={{TrafficSim}: Learning to simulate realistic multi-agent behaviors},
  author={Suo, Simon and Regalado, Sebastian and Casas, Sergio and Urtasun, Raquel},
  booktitle=CVPR,
  pages={10400--10409},
  year={2021}
}

@inproceedings{bergamini2021simnet,
  title={{SimNet}: Learning reactive self-driving simulations from real-world observations},
  author={Bergamini, Luca and Ye, Yawei and Scheel, Oliver and Chen, Long and Hu, Chih and Del Pero, Luca and Osi{\'n}ski, B{\l}a{\.z}ej and Grimmett, Hugo and Ondruska, Peter},
  booktitle=ICRA,
  pages={5119--5125},
  year={2021},
}

@article{xu2022bits,
  title={{BITS}: Bi-level imitation for traffic simulation},
  author={Xu, Danfei and Chen, Yuxiao and Ivanovic, Boris and Pavone, Marco},
  journal={arXiv preprint arXiv:2208.12403},
  year={2022}
}

@article{peng2025diffusion,
  title={Diffusion models for intelligent transportation systems: A survey},
  author={Peng, Mingxing and Chen, Kehua and Guo, Xusen and Zhang, Qiming and Zhong, Hui and Zhu, Meixin and Yang, Hai},
  journal={IEEE Transactions on Intelligent Transportation Systems},
  year={2025},
}

@inproceedings{xie2024advdiffuser,
  title={{AdvDiffuser}: Generating Adversarial Safety-Critical Driving Scenarios via Guided Diffusion},
  author={Xie, Yuting and Guo, Xianda and Wang, Cong and Liu, Kunhua and Chen, Long},
  booktitle=IROS,
  pages={9983--9989},
  year={2024},
}

@inproceedings{zhong2023guided,
  title={Guided conditional diffusion for controllable traffic simulation},
  author={Zhong, Ziyuan and Rempe, Davis and Xu, Danfei and Chen, Yuxiao and Veer, Sushant and Che, Tong and Ray, Baishakhi and Pavone, Marco},
  booktitle=ICRA,
  pages={3560--3566},
  year={2023},
}

@inproceedings{jiang2023motiondiffuser,
  title={{MotionDiffuser}: Controllable multi-agent motion prediction using diffusion},
  author={Jiang, Chiyu and Cornman, Andre and Park, Cheolho and Sapp, Benjamin and Zhou, Yin and Anguelov, Dragomir and others},
  booktitle=CVPR,
  pages={9644--9653},
  year={2023}
}

@inproceedings{zhong2023language,
  title={Language-guided traffic simulation via scene-level diffusion},
  author={Zhong, Ziyuan and Rempe, Davis and Chen, Yuxiao and Ivanovic, Boris and Cao, Yulong and Xu, Danfei and Pavone, Marco and Ray, Baishakhi},
  booktitle=CORL,
  pages={144--177},
  year={2023},
}

@article{peng2025safety,
  title={Safety-Critical Traffic Simulation with Guided Latent Diffusion Model},
  author={Peng, Mingxing and Yao, Ruoyu and Guo, Xusen and Xie, Yuting and Chen, Xianda and Ma, Jun},
  journal={arXiv preprint arXiv:2505.00515},
  year={2025}
}

@article{peng2025ld,
  title={{LD-Scene}: Llm-guided diffusion for controllable generation of adversarial safety-critical driving scenarios},
  author={Peng, Mingxing and Xie, Yuting and Guo, Xusen and Yao, Ruoyu and Yang, Hai and Ma, Jun},
  journal={arXiv preprint arXiv:2505.11247},
  year={2025}
}

@article{philion2023trajeglish,
  title={Trajeglish: Traffic modeling as next-token prediction},
  author={Philion, Jonah and Peng, Xue Bin and Fidler, Sanja},
  journal={arXiv preprint arXiv:2312.04535},
  year={2023}
}

@inproceedings{seff2023motionlm,
  title={{MotionLM}: Multi-agent motion forecasting as language modeling},
  author={Seff, Ari and Cera, Brian and Chen, Dian and Ng, Mason and Zhou, Aurick and Nayakanti, Nigamaa and Refaat, Khaled S and Al-Rfou, Rami and Sapp, Benjamin},
  booktitle=ICCV,
  pages={8579--8590},
  year={2023}
}

@inproceedings{wu2024smart,
  title={{SMART}: Scalable multi-agent real-time motion generation via next-token prediction},
  author={Wu, Wei and Feng, Xiaoxin and Gao, Ziyan and Kan, Yuheng},
  booktitle=NIPS,
  volume={37},
  pages={114048--114071},
  year={2024}
}

@inproceedings{zhang2025closed,
  title={Closed-loop supervised fine-tuning of tokenized traffic models},
  author={Zhang, Zhejun and Karkus, Peter and Igl, Maximilian and Ding, Wenhao and Chen, Yuxiao and Ivanovic, Boris and Pavone, Marco},
  booktitle=CVPR,
  pages={5422--5432},
  year={2025}
}

@article{zhou2025decoupled,
  title={Decoupled diffusion sparks adaptive scene generation},
  author={Zhou, Yunsong and Ye, Naisheng and Ljungbergh, William and Li, Tianyu and Yang, Jiazhi and Yang, Zetong and Zhu, Hongzi and Petersson, Christoffer and Li, Hongyang},
  journal={arXiv preprint arXiv:2504.10485},
  year={2025}
}

@article{chen2024diffusion,
  title={{Diffusion Forcing}: Next-token prediction meets full-sequence diffusion},
  author={Chen, Boyuan and Mart{\'\i} Mons{\'o}, Diego and Du, Yilun and Simchowitz, Max and Tedrake, Russ and Sitzmann, Vincent},
  journal={Advances in Neural Information Processing Systems},
  volume={37},
  pages={24081--24125},
  year={2024}
}

@inproceedings{peebles2023scalable,
  title={Scalable diffusion models with transformers},
  author={Peebles, William and Xie, Saining},
  booktitle=ICCV,
  pages={4195--4205},
  year={2023}
}

@article{yao2024calmm,
  title={{CALMM-Drive}: Confidence-Aware Autonomous Driving with Large Multimodal Model},
  author={Yao, Ruoyu and Wang, Yubin and Liu, Haichao and Yang, Rui and Peng, Zengqi and Zhu, Lei and Ma, Jun},
  journal={arXiv preprint arXiv:2412.04209},
  year={2024}
}

@article{lin2002divergence,
  title={Divergence measures based on the Shannon entropy},
  author={Lin, Jianhua},
  journal={IEEE Transactions on Information Theory},
  volume={37},
  number={1},
  pages={145--151},
  year={2002},
}

@inproceedings{heusel2017gans,
  title={Gans trained by a two time-scale update rule converge to a local nash equilibrium},
  author={Heusel, Martin and Ramsauer, Hubert and Unterthiner, Thomas and Nessler, Bernhard and Hochreiter, Sepp},
  booktitle=NIPS,
  volume={30},
  year={2017}
}

@article{shannon1948mathematical,
  title={A mathematical theory of communication},
  author={Shannon, Claude E},
  journal={The Bell System Technical Journal},
  volume={27},
  number={3},
  pages={379--423},
  year={1948},
}

@inproceedings{dauner2023parting,
  title={Parting with misconceptions about learning-based vehicle motion planning},
  author={Dauner, Daniel and Hallgarten, Marcel and Geiger, Andreas and Chitta, Kashyap},
  booktitle=CORL,
  pages={1268--1281},
  year={2023},
}

@inproceedings{ho2020denoising,
  title={Denoising diffusion probabilistic models},
  author={Ho, Jonathan and Jain, Ajay and Abbeel, Pieter},
  booktitle=NIPS,
  pages={6840--6851},
  year={2020}
}

@inproceedings{cheng2024rethinking,
  title={Rethinking imitation-based planners for autonomous driving},
  author={Cheng, Jie and Chen, Yingbing and Mei, Xiaodong and Yang, Bowen and Li, Bo and Liu, Ming},
  booktitle=ICRA,
  pages={14123--14130},
  year={2024},
}

@article{helbing1998generalized,
  title={Generalized force model of traffic dynamics},
  author={Helbing, Dirk and Tilch, Benno},
  journal={Physical review E},
  volume={58},
  number={1},
  pages={133},
  year={1998},
}

@inproceedings{scheel2022urban,
  title={{Urban Driver}: Learning to drive from real-world demonstrations using policy gradients},
  author={Scheel, Oliver and Bergamini, Luca and Wolczyk, Maciej and Osi{\'n}ski, B{\l}a{\.z}ej and Ondruska, Peter},
  booktitle=CORL,
  pages={718--728},
  year={2022},
}

@inproceedings{zhengdiffusion,
  title={Diffusion-Based Planning for Autonomous Driving with Flexible Guidance},
  author={Zheng, Yinan and Liang, Ruiming and Zheng, Kexin and Zheng, Jinliang and Mao, Liyuan and Li, Jianxiong and Gu, Weihao and Ai, Rui and Li, Shengbo Eben and Zhan, Xianyuan and others},
  booktitle=ICLR,
  year={2025}
}

@inproceedings{hallgarten2023prediction,
  title={From prediction to planning with goal conditioned lane graph traversals},
  author={Hallgarten, Marcel and Stoll, Martin and Zell, Andreas},
  booktitle={the Proceedings of 2023 IEEE 26th International Conference on Intelligent Transportation Systems},
  pages={951--958},
  year={2023},
}

\end{document}